\documentclass{article}

\PassOptionsToPackage{numbers, compress}{natbib}

 \usepackage[preprint]{neurips_2026}

\usepackage[utf8]{inputenc} 
\usepackage[T1]{fontenc}    
\usepackage{hyperref}       
\usepackage{url}            
\usepackage{booktabs}       
\usepackage{amsmath}        
\usepackage{amssymb}        
\usepackage{amsfonts}       
\usepackage{nicefrac}       
\usepackage{microtype}      
\usepackage{xcolor}         
\usepackage{graphicx}       
\usepackage{algorithm}      
\usepackage{algorithmic}    

\title{A Counterfactual Graph Approach to Oppositional Narrative Analysis}

\author{%
  Diego Revilla\thanks{These authors contributed equally to this work.}\,\,\,\thanks{Corresponding authors.} \\
  Faculty of Engineering, University of Deusto, Bilbao, Spain \\
  HiTZ Center -- Ixa, University of the Basque Country (UPV/EHU), Donostia, Spain \\
  \texttt{diego.r@opendeusto.es} \\
  \And
  Mart\'in Fern\'andez-de-Retana\footnotemark[1] \\
  Faculty of Engineering, University of Deusto\\
  Bilbao, Spain \\
  \texttt{martin.f@opendeusto.es}\\
  \And
  Lingfeng Chen\footnotemark[1] \\
  Faculty of Engineering, University of Deusto\\
  Bilbao, Spain \\
  \texttt{lingfeng.chen@opendeusto.es}\\
  \AND
  Aritz Bilbao-Jayo\thanks{These authors jointly supervised this work.} \\
  Faculty of Engineering, University of Deusto\\
  Bilbao, Spain \\
  \texttt{aritzbilbao@deusto.es} \\
  \And
  Miguel Fernandez-de-Retana\footnotemark[3] \,\footnotemark[2] \\
  Faculty of Engineering, University of Deusto\\
  Bilbao, Spain \\
  \texttt{m.fernandezderetana@deusto.es} \\
}

\begin{document}

\maketitle

\begin{abstract}

Distinguishing legitimate critical speech from coordinated conspiracy narratives is a pressing challenge for the integrity of the online information ecosystem, yet prevailing approaches rely on labeled entity roles and rigid predefined schemas that fail to capture the diffuse and context-dependent nature of participation in such narratives. We propose a modular framework that extracts an entity-interaction graph directly from narratives and reasons over it for classification, without depending on annotated node-level labels. Our pipeline enumerates candidate entity spans from a domain-adapted BERT encoder and learns a bipartite graph structure, which is then classified with a Heterogeneous Graph Transformer. To make predictions interpretable, we cast the graph as a hypergraph and apply a counterfactual distillation procedure that estimates the Individual Treatment Effect of each node, pruning extraneous entities to recover a minimal predictive subgraph. On the standard PAN Oppositional Narratives benchmark, our model achieves a macro $F_1$-score of 0.93 and an MCC of 0.84, reaching \textit{state-of-the-art} performance with roughly one-third of the parameters of the next best system, while pruning the graphs by 75.2\% on average without altering predictions. Our results demonstrate that graph-based representations coupled with counterfactual inference offer both strong predictive accuracy and transparency into the entities driving conspiratorial discourse.
\end{abstract}

\section{Introduction}

Conspiracy theories pose a significant threat to society on multiple levels. They can be deliberately exploited to manipulate public opinion and exert pressure on public institutions, thereby influencing their decision-making processes \citep{demoCons, ConspiracyPsychology}. During the COVID-19 pandemic, conspiracy narratives and targeted attacks circulating on social media became a major risk factor that heightened public distrust in institutions \citep{COVID19Conspiracy}. The automated and timely identification of such coordinated narratives and potential aggregated attacks is therefore of critical importance.

Distinguishing critical speech from conspiratorial attacks within a narrative is a complex challenge in natural language processing. The task typically requires uncovering the entities in the input text together with the relational structure that defines the roles they play in such attacks. This is rarely straightforward: the entities actually involved are often introduced through subtle contextual cues and terms used with concealed intent. Such subtlety complicates automation and, in many cases, leads annotators to introduce personal biases or to fail to resolve references correctly \citep{geva2019we, sap2022annotators}.

These difficulties are not merely a matter of accuracy. Systems that moderate or triage public discourse act on speech, and a classifier that cannot expose \emph{why} a message was flagged offers no basis for appeal, audit, or oversight. Opaque detectors trained on rigid annotation schemas therefore inherit both the biases of their annotators and an inability to justify their decisions, which makes deployment in civic and institutional settings precisely untrustworthy. We argue that models operating in this domain should expose the intermediate structure on which their decisions rest.

Our main contribution lies in extracting a graph representation from narrative text and subsequently distilling this graph to identify the primary entities driving a conspiracy narrative. Unlike approaches that depend on predefined role labels, which often introduce edge cases and fail to capture the inherently continuous and diffuse nature of participation \citep{conflicts_texts}, our method avoids rigid role assignments. Instead, we let the model learn to identify, classify, and filter textual spans to construct a coherent and contextually appropriate representation of the discourse for classification. In summary, the key contributions of this work\footnote{The implementation of the model and the code to replicate the experiments can be found on \href{https://github.com/Lingfeng555/Causal-Entity-Framing-Graph/}{GitHub}.} are:

\begin{itemize}
    \item \textbf{Graph Representation:} A pipeline that extracts entity-interaction graphs, modeling structural interactions without predefined role schemas.
    \item \textbf{Minimal Graph Distillation:} A distillation procedure that estimates the Individual Treatment Effect (ITE) of nodes to prune extraneous nodes and recover a minimal predictive subgraph using counterfactual evaluation of the subgraphs.
    \item \textbf{Text Classification:} A state-of-the-art classifier reaching an $F_1$-score of 0.93, demonstrating the adequacy of graphs for detecting conspiracy vectors.
\end{itemize}

\section{Related Work}

\textbf{Entity Framing:}
The notion of assigning roles to entities in text traces back to frame semantics and the FrameNet project \citep{baker1998berkeley}, which characterized how participants are framed within a predicate's semantic structure. This intuition was later operationalized as Semantic Role Labeling, where the goal is to identify the arguments of each predicate and label their roles, as standardized in the CoNLL-2005 shared task \citep{carreras2005introduction}.

The first adaptations of these ideas to narrative analysis in the NLP community framed entities according to classic narrative archetypes. The Narrative Policy Framework \citep{NPF} applied content analysis to identify the roles of hero, villain, and victim in policy discourse. Concurrently, schemes for analyzing persuasion were proposed, such as the detection of propaganda techniques in news articles \citep{da2019fine}. Building on this line of work, subsequent approaches leveraged role dictionaries and sentiment analysis to automate role detection \citep{NLTKRolesDetection}, and the task was eventually generalized to entity framing, combining the identification of characters with the classification of their assigned roles. Currently, the predominant technique in the state of the art consists of fine-tuning a foundational LLM for this task \citep{SOTAEntityFraming}.



\textbf{Inherent Graph Representation:} 
Many problems and real-world scenarios can be naturally represented as graphs, such as RNA structure \citep{RNAGNN} and protein-protein interaction \citep{xu2024graphneuralnetworksproteinprotein}. This structural perspective has motivated the development of Graph Neural Networks (GNNs) \citep{FirstGNN}. However, certain tasks require the explicit modeling of causal relationships within the underlying context, thereby motivating the integration of causal inference principles into graph-based methods. This line of work has given rise to Causal Graph Neural Networks (Causal GNNs) \citep{FirstCausalGNN, InstrumentalCausalInference}. Subsequently, causal discovery methods were further enhanced through GNN-based architectures that learn distributions over the space of possible enumerations of subgraphs \citep{CausalDiscovery}.

In the field of natural language processing, several efforts have been made to transform human-readable text into graph-based representations. The main goal is to improve the performance of current state-of-the-art models on tasks such as text classification, where subcontextual information can be captured more effectively in graphs \citep{ConectConectingDots}. Within this line of research, linguistic approaches use large language models (LLMs) to summarize information and convert it into nodes, which are then connected through semantic relationships \citep{TextToGraphLinguisticApproach}. Graph-based text classification can also be improved with contrastive learning and data augmentation, where augmented graphs are generated from the text to improve generalization performance \citep{TextLevelAugmentedGraph}. Furthermore, unsupervised methods based on cyclic training have been proposed, especially for smaller datasets, where the high cost of data collection makes fully supervised learning less feasible \citep{UnsupervisedLearning}.

\textbf{Graph Interpretation Problem:} 
Evaluating and interpreting graphical representations of real-world information to determine their quality, as well as assessing the extent of information preserved or lost, remains a challenging task. Focusing on the evaluation of a graphical representation, HL-HGAT (Hodge-Laplacian Heterogeneous Graph Attention Network) \citep{HLGraphEvaluation} is a graph neural network that extends GNNs to handle not only nodes and edges, but also higher-dimensional simplices (e.g., triangles). In terms of interpretability and explainability, several traditional techniques have been adapted to graph representations. FactGraph is a framework that evaluates the factual consistency of a graph representation of a summary (or text) with respect to the source document \citep{FactGraph}.

From the perspective of explainability, several adaptations of classical machine learning approaches to graphs have been presented in recent years, such as gradient-based methods and counterfactual explanations \citep{GraphXAISurvey}. CF-GNNExplainer is a counterfactual explanation method based on distortion matrices, which generates perturbations in the input graph to produce counterfactual examples that reveal how the graph structure influences the model's behavior \citep{CF-GNNExplainer}.

\section{Methodology}

\begin{figure}[t]
    \centering
    \includegraphics[width=\linewidth]{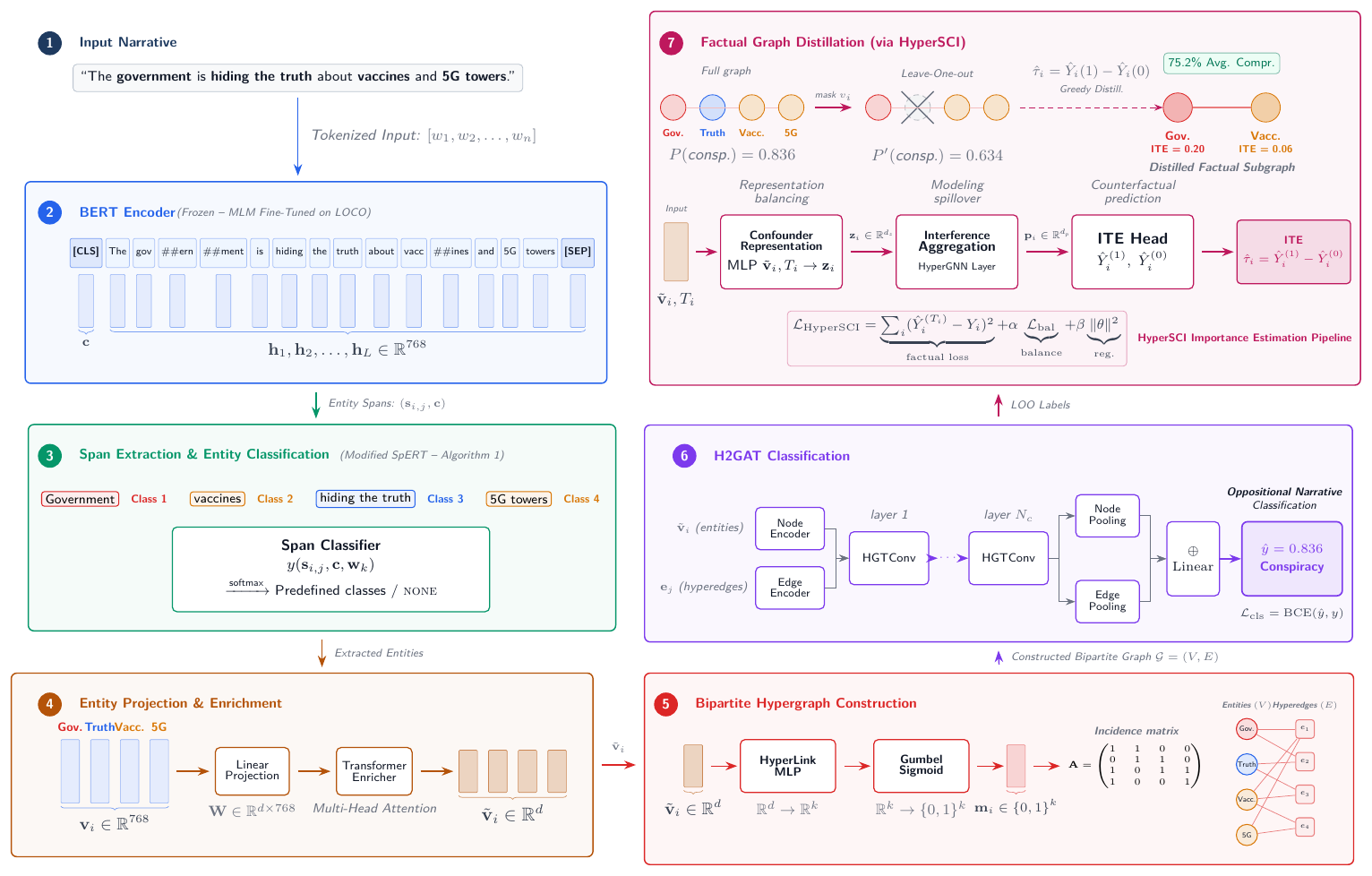}
    \caption{Overview of the proposed methodology. Stage~1 defines the input, which in this case consists of oppositional narratives. Stage~2 generates tokens and contextual embeddings using BERT. Stage~3 extracts the entities $V$ from the tokens and embeddings by applying Algorithm~\ref{alg:hypergraph_vertices} (Appendix~\ref{sec:appendix}). Stage~4 produces enriched low-dimensional representations of the entities. Stage~5 constructs the incidence matrix $\mathbf{A}$. Stage~6 performs graph classification. The final stage presents the counterfactual analysis. The key point is that every classification decision is routed through an explicit entity-interaction graph, which makes the counterfactual distillation of the last stage possible. Entity labels are abbreviated for readability; actual nodes are token spans (cf.\ Figure~\ref{fig:example_prediction}).}
    \label{fig:architecture}
\end{figure}

Our approach is inspired by SpERT \citep{SPERT}, which proposes an iterative algorithm for joint entity extraction and classification with a strong reliance on accurately labeled entity spans. While SpERT demonstrates strong performance, it depends heavily on high-quality annotations. Our proposal extends this paradigm by allowing the model to learn to assign different attention classes to the resulting nodes and by incorporating interpretability into the model's predictions. Each stage of the proposed methodology is illustrated in Figure~\ref{fig:architecture}.

\subsection{Text Encoder \& SpERT}

SpERT is based on a BERT \citep{BERT} encoder used to extract word embeddings, which are later employed to extract the candidate spans. Because our architecture must jointly maximize classification accuracy and counterfactual interpretability, we use two separate fine-tuning strategies for the encoder depending on the preferred objective:

\begin{itemize}
    \item \textbf{Classification Pre-Training:} To improve the encoder for our concrete task, we fine-tuned it on the LOCO dataset \citep{LOCO} for conspiracy classification. It was subsequently trained on our target dataset and evaluated on a holdout test set to assess generalization.

    \item \textbf{Domain-Adaptive Pre-Training:} For the causal inference and interpretability phase, we subject the base model (BERT) to an unsupervised fine-tuning process on the domain-specific corpus using the Masked Language Modeling (MLM) task instead of the classification task mentioned above. Due to BERT's attention mechanism, information about an entity tends to be distributed among the attended tokens. If the encoder were trained dynamically alongside the model, it could learn to ``hide'' entity information in the neighboring context in order to maximize classification performance \citep{aiayn, he2021deberta}, thereby invalidating the semantic information carried by each token.
\end{itemize}

After fine-tuning BERT, we keep the encoder frozen for the rest of the pipeline.

\subsection{Span Extraction \& Entity Classification}
\label{sec:span}

One key assumption of this work is that if a span belongs to a given category, then all of its subspans also belong to the same category. However, the converse does not necessarily hold: if a span does not belong to a category, this does not imply that its subspans are excluded from that category. For example, if \textit{``President Obama''} is annotated as an entity, then \textit{``President''} is also considered an entity in the corresponding discourse under this assumption. In contrast, a longer span such as \textit{``The President Obama from the White House''} may not be annotated as an entity, even though it contains a valid entity as a subspan.

Based on this assumption, we propose a span-framing procedure to extract all possible entity candidates from the input narrative (see Algorithm~\ref{alg:hypergraph_vertices} in Appendix~\ref{sec:appendix}). The algorithm enumerates spans greedily up to a maximum width of $w_{max}$. Each candidate span of width $k$ starting at position $i$ is represented by concatenating three vectors: the sentence-level \texttt{[CLS]} embedding, the component-wise maximum pooling of the $k$ constituent embeddings, and a learned span-width embedding. This representation is passed to a linear span classifier that assigns one of $N$ predefined, fixed entity types or a null \emph{no-span} class.

\subsection{Entity Projection \& Enrichment}

Once all entities are extracted, each entity is linearly projected into a lower-dimensional embedding space in order to reduce the latent dimensionality and alleviate overfitting. These projected embeddings are then jointly enriched using a lightweight Transformer encoder, which models contextual interactions among all extracted entities.

\subsection{Bipartite Graph Construction}

We train the model to generate a bipartite graph $G = (V, E, L)$, where $V = \{v_{1}, \dots, v_{d}\}$ is the set of $d$ extracted entities, $E = \{e_{1}, \dots, e_{k}\}$ is a predefined set of $k$ relation nodes that act as relational factors or correlation nodes between entities, and $L$ denotes the bipartite edges. For each entity $v_{i} \in V$, a multilayer perceptron produces a binary adjacency vector indicating its connections to the relation nodes in $E$. Discrete relation membership is obtained via a Gumbel-Sigmoid estimator \citep{gumbel} followed by hard thresholding, which enables end-to-end training through the discrete selection. These connections are collected in a binary incidence matrix $\mathbf{A} \in \{0, 1\}^{d \times k}$, where $\mathbf{A}_{i,j} = 1$ if entity $v_{i}$ is linked to relation $e_{j}$.

To enable our downstream causal distillation, we formally project this bipartite structure into an undirected hypergraph $\mathcal{H} = (\mathcal{V}, \mathcal{E})$. The hypergraph node set is the set of extracted entities, $\mathcal{V} = V$, and each relation node $e_{j} \in E$ induces a single hyperedge $\varepsilon_{j} \in \mathcal{E}$ defined as the subset of entities connected to it:
\begin{equation}
\varepsilon_{j} = \{v_{i} \in \mathcal{V} \mid \mathbf{A}_{i,j} = 1\}
\end{equation}
The raw hyperedge set is thus $\mathcal{E} = \{\varepsilon_{1}, \dots, \varepsilon_{k}\}$. To ensure structural validity and handle edge cases (e.g., relations with zero or one connected entity), we restrict the final hyperedge set to those that capture genuine structural interactions:
\begin{equation}
\mathcal{E}_{valid} = \{\varepsilon_{j} \in \mathcal{E} \mid |\varepsilon_{j}| \ge 2\}
\end{equation}
This formalization lets us treat the entity-interaction space purely as a hypergraph for the subsequent HyperSCI analysis.

\subsection{Heterogeneous Graph Classification}

To enable the model to learn meaningful graph-based representations of the narrative text, we employ a Heterogeneous Graph Transformer (HGT) \citep{H2GAT} for graph-level classification. HGT is particularly suitable for modeling narrative graphs containing correlated entities and coordinated interactions, as it explicitly accounts for structural and semantic heterogeneity through class-aware relational attention. By assigning node- and edge-type-dependent projection matrices, HGT captures role-specific influence patterns and interaction asymmetries while maintaining parameter sharing for efficient generalization. These properties make HGT especially effective for identifying implicit relational motifs, modeling higher-order dependencies, and detecting orchestrated or aggregated attacks in complex narrative graphs.

The HGT operates over the constructed bipartite graph and is trained under supervision at the narrative level. Specifically, each graph corresponding to a full text is assigned a ground-truth global label (\textit{critical} or \textit{conspiracy}), and the model is trained to predict this label, thereby learning graph representations that capture the structural discriminative patterns for narrative classification.

\begin{table}[t]
\caption{Decomposition of the total number of model parameters by component. Over 99.8\% of the parameters live in the frozen text encoder: the entire graph construction and reasoning machinery adds fewer than 200K trainable parameters.}
\label{tab:param_breakdown}
\centering
\small
\begin{tabular}{lrr}
\toprule
\textbf{Component} & \textbf{Parameters} & \textbf{Percentage (\%)} \\
\midrule
BERT (frozen) & 109,483,778 & 99.841 \\
Enricher & 2,240 & 0.002 \\
HyperLink classifier & 759 & 0.001 \\
Hypergraph classifier & 126,745 & 0.116 \\
Entity projector & 24,592 & 0.022 \\
Span classifier & 11,525 & 0.011 \\
Width embeddings & 4,608 & 0.004 \\
Type embeddings & 3,840 & 0.004 \\
\midrule
\textbf{Total} & \textbf{109,658,087} & \textbf{100.00} \\
\bottomrule
\end{tabular}
\end{table}

Although the overall architecture is complex, the vast majority of its parameters are concentrated in the frozen text encoder. The graph construction and reasoning components introduce fewer than 200K additional parameters, as detailed in Table~\ref{tab:param_breakdown}.

\subsection{Counterfactual Graph Distillation}

To identify the nodes that drive the classification decision for each graph, we implement the HyperSCI model \citep{hyperSCI}, which estimates the Individual Treatment Effect (ITE) as an indicator of node influence \citep{zhang2025individualtreatmenteffectprediction} while accounting for the interactions between nodes through hyperedges, i.e., settings in which the causal effects of different units are not independent. In this context, the treatment refers to the active presence of a specific node within the text. We therefore attempt to measure the isolated impact that a single entity has on the model's final classification. Although this effect can be computed exactly by masking each node and re-running the classifier, HyperSCI amortizes the estimate into a single forward pass per graph and explicitly models spillover between entities that share a hyperedge.

\paragraph{Confounder Representation Learning.}
In observational studies, a confounder \citep{confounder} is a variable that distorts the estimated relationship between treatment and outcome by being independently associated with both. To control for this, HyperSCI uses an MLP to project the node features $X$ into a latent space $Z$ intended to capture all potential confounders. To keep the confounder distribution invariant across treatment and control nodes after this projection, a regularization term based on the Wasserstein distance \citep{Chewi2025} penalizes the discrepancy between the two distributions:
\begin{equation}
\mathcal{L}_{was} = \text{Wasserstein}(Z_{treatment}, Z_{control})
\end{equation}

\paragraph{Interference Modeling.}
The interference module propagates the confounder representations $Z$ through the hypergraph $\mathcal{H}$. For a node $i$, its interference representation $P_i$ aggregates spillover information, i.e., information propagated from neighboring nodes:
\begin{equation}
H^{(l+1)} = \sigma(\text{HyperGNN}(H^{(l)}, \mathcal{H}, T))
\end{equation}
where $T$ weights the propagation, letting the model capture how the presence or absence of neighboring entities alters the predictive context of node $i$. This allows us to estimate the ITE under spillover effects.

\paragraph{ITE Prediction.}
Finally, we concatenate the confounder representation $z_i$ and the interference representation $P_i$ and feed them to two prediction heads that output $\hat{Y}_i(1)$, the predicted outcome when node $i$ is present, and $\hat{Y}_i(0)$, the predicted outcome when it is absent. The ITE is then computed as:
\begin{equation}
\hat{\tau}_i = \hat{Y}_i(1) - \hat{Y}_i(0)
\end{equation}

\paragraph{Leave-One-Out Supervision.}
Since we lack ground-truth counterfactual labels, we construct a synthetic dataset using a Leave-One-Out (LOO) scheme. For each text $d$ in the training set, we define
\begin{equation*}
Y_i(t) = P\!\left(\text{conspiracy} \mid \mathcal{H}_{d\setminus (1-t)\{i\}}\right), \quad t \in \{0,1\}
\end{equation*}
where $\mathcal{H}_d$ is the complete hypergraph of document $d$, and $\mathcal{H}_{d\setminus\{i\}}$ denotes the same hypergraph with the features of node $i$ set to zero. Note that $Y_i(1)$ is identical for all nodes in a given document (the base probability), whereas $Y_i(0)$ is specific to each node. The reference effect $\text{ITE}_{ref}$ is then defined as:
\begin{align*}
    \tau_i &= Y_i(1) - Y_i(0)\\
    &= P\!\left(\text{conspiracy} \mid \mathcal{H}_d\right) - P\!\left(\text{conspiracy} \mid \mathcal{H}_{d\setminus \{i\}}\right)
\end{align*}
The final objective function combines the mean squared error (MSE) with a regularization term:
\begin{equation}
\mathcal{L} = \sum_{i} \left(\hat{y}_{i,\text{factual}} - y_{i,\text{true}}\right)^2 + \alpha \cdot \mathcal{L}_{was}
\end{equation}
where $y_{i,\text{true}}$ denotes the observed factual outcome, $\hat{y}_{i,\text{factual}}$ represents the prediction produced by the HyperSCI model for the factual instance, and $\alpha$ is a hyperparameter that controls the contribution of the Wasserstein regularization term $\mathcal{L}_{was}$.

Once the ITEs have been computed, we recover the core predictive nodes through a greedy search: nodes are sorted by descending ITE and progressively unmasked from the graph until the model produces a prediction contrary to the original one.

\section{Results}

\subsection{Dataset}

We test the proposed architecture on the dataset from PAN Oppositional Narratives \citep{korenvcic2024overview}, which comes from the XAI-DisInfodemic corpus \citep{Koren_i__2024} and consists of 5,000 English Telegram messages containing dissenting and non-conventional opinions about the COVID-19 pandemic. These messages are divided into two categories: those suggesting the existence of a conspiracy and those that criticize mainstream views on the COVID-19 pandemic without implying the existence of a conspiracy.

The training split contains 4,000 samples, encompassing 1,379 conspiratorial and 2,621 critical texts. This class distribution is closely mirrored in the 1,000-sample test set, which consists of 345 conspiratorial and 655 critical instances, ensuring consistent class proportions across evaluation.

A key limitation of this dataset is its relatively small size for training classification tasks. To mitigate this issue, we make use of the LOCO (Language Of Conspiracy) dataset \citep{LOCO}, an 88-million-token, open-source corpus featuring 23,937 conspiracy and 72,806 topic-matched mainstream documents from 150 websites. The BERT submodel was fine-tuned on this considerably larger corpus and then frozen, giving the classifier access to linguistic representations already adapted to the domain.

\subsection{Classification Performance}

\begin{table}[t]
    \caption{Classification results of our model compared to the PAN benchmark \citep{korenvcic2024overview}. Our model outperforms IUCL \citep{iucl}, AI\_Fusion \citep{ayele2024panoverview} and SINAI \citep{sinai} on every metric while using roughly one-third of the parameters of the runner-up. Fine-tuned BERT is adapted to the specific context of the target class, whereas BERT+LOCO is a fine-tuned version trained on a larger dataset, enabling broader domain adaptation across several types of conspiracy-related texts.}
    \label{tab:task1_english_results}
    \small
    \centering
    \setlength{\tabcolsep}{6pt}
    \begin{tabular}{lcccc}
        \toprule
        Team & MCC$^{\uparrow}$ & $F_1$-Macro$^{\uparrow}$ & $F_1$-Cons.$^{\uparrow}$ & $F_1$-Crit.$^{\uparrow}$ \\
        \midrule
        Our model                & \textbf{0.840} & \textbf{0.930} & \textbf{0.895} & \textbf{0.945} \\
        IUCL                     & 0.838 & 0.919 & 0.894 & 0.944 \\
        AI\_Fusion               & 0.830 & 0.914 & 0.886 & 0.942 \\
        SINAI                    & 0.829 & 0.914 & 0.888 & 0.941 \\
        \midrule
        Fine-tuned BERT          & 0.796 & 0.897 & 0.863 & 0.931 \\
        BERT+LOCO                & 0.655 & 0.844 & 0.787 & 0.901 \\
        \bottomrule
    \end{tabular}
\end{table}

The proposed system achieves a macro $F_1$-score of 0.93, misclassifying only 72 of the 1{,}000 samples in the test set. As shown in Table~\ref{tab:task1_english_results}, when compared against the official task results, our model outperforms all other approaches and reaches a Matthews Correlation Coefficient (MCC) of 0.840. The approach is also highly parameter-efficient: as detailed in Table~\ref{tab:param_breakdown}, it operates with approximately 109M parameters, roughly one-third of the parameter count of the second-place solution, which relies on a heavier DeBERTa-large model (304M parameters).

\subsection{Counterfactual Interpretability}

We evaluate node criticality using two criteria:

\begin{itemize}
    \item \textbf{Precision in Estimation of Heterogeneous Effect (PEHE)} \citep{Hill01012011}: an error metric reflecting how accurately our model estimates the individual-level treatment effects (ITEs). Lower PEHE values indicate higher precision in isolating the true causal impact.

    \item \textbf{Average Treatment Effect (ATE)}: the mean impact of the treatment on the classification outcome across all nodes.
\end{itemize}

Our model achieves a PEHE of 0.2994 and an ATE of 0.0218. These results indicate that the general estimation error remains non-negligible, while the relatively low ATE suggests that the average counterfactual differences across individual units are modest.

\begin{figure}[t]
    \centering
    \includegraphics[width=0.53\linewidth]{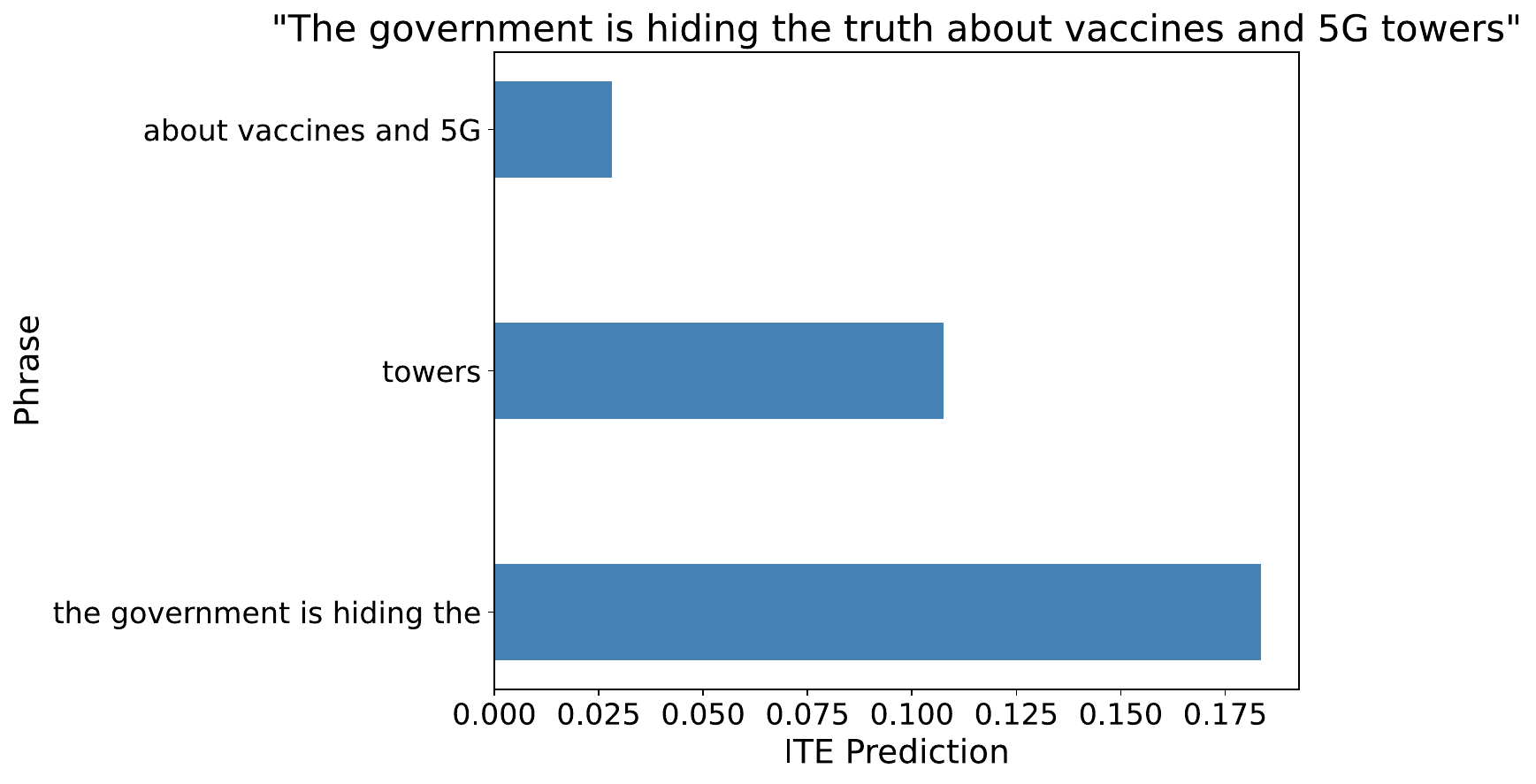}
    \caption{Per-entity ITE contributions for an oppositional narrative text correctly classified as conspiracy, with a predicted probability of 0.836. A small number of entities account for most of the positive contribution which allows the graph to be pruned aggressively without flipping the prediction.}
    \label{fig:example_prediction}
\end{figure}

While the architecture captures several of the most critical relationships, the PEHE indicates room to improve the accuracy of individual effect estimates, and the ATE indicates that counterfactual impact is distributed granularly across nodes rather than concentrated in a few dominant entities. On the basis of these metrics, we regard the counterfactual component as a proof of concept with room for improvement, and we report it as such. To analyze interpretability, ITEs were computed for selected text examples. One such example is illustrated in Figure~\ref{fig:example_prediction}.

\subsection{Minimal Subgraph Pruning}

To identify the minimal subgraph, we quantified the extent to which the entity graph can be compressed while preserving correct predictions. The \textit{compression rate} (CR) is defined as:
\begin{equation*}
\text{CR} = \left(1 - \frac{\lvert V_{\text{remaining}} \rvert}{\lvert V \rvert}\right) \times 100
\end{equation*}
On average, the model requires only a small fraction of entities to make a correct decision, suggesting effective counterfactual attribution. In future work this should be tested with a larger number of classes in order to isolate the specific predictive value of the graph structures: because the final classification head uses a sigmoid activation, the model naturally reverts to the dataset's baseline distribution when graph signals are fully compressed.

Further insights into the model's counterfactual attribution can be observed through the analysis of minimal subgraph compression rates. The system achieves an overall average compression rate of 75.2\%, with a standard deviation of $\sigma = 7.8$ and a maximum of 96.2\%. When analyzing performance by category, the mean compression rate is 72.8\% for critical texts and 79.6\% for conspiratorial texts. These figures demonstrate that the model is capable of formulating accurate predictions relying on only a small fraction of the entities present in the graph.

\section{Ablations}

An ablation study was conducted to assess the contribution of each module to classification performance; for reference, the full model achieves an $F_1$-score of approximately $0.93$. Removing the enrichment module that relies on the BERT model fine-tuned for MLM decreased the $F_1$-score to $0.86$, a relative decrease of $7.53\%$.

Removing the entity classifier had a far stronger effect, causing the model weights to collapse during training: in this ablation all candidate entities had to be treated as valid, since the original approach uses the classifier to filter out irrelevant candidates, and the $F_1$-score fell to $0.5256$ ($43.48\%$ relative decrease). Finally, removing the Gumbel-Sigmoid module and forcing a fully connected graph instead of learning its structure reduced the $F_1$-score to $0.3293$, a relative decrease of $64.59\%$. Each module therefore contributes meaningfully to the overall performance of the counterfactual inference model; all results are reported in Table~\ref{tab:ablation_result}.

\begin{table}[t]
    \caption{Ablation study results and performance decrease. Learning the graph structure with Gumbel-Sigmoid is the single most important component: forcing a fully connected graph costs almost two-thirds of the model's performance.}
    \label{tab:ablation_result}
    \small
    \centering
    \setlength{\tabcolsep}{6pt}
    \begin{tabular}{lcc}
    \toprule
    \textbf{Ablation Approach} & \textbf{$F_1$-Score} & \textbf{Relative Decrease} \\
    \midrule
    Original model            & $0.9300$ & -- \\
    No enrichment module      & $0.8600$ & $7.53\%$ \\
    No entity classifier      & $0.5256$ & $43.48\%$ \\
    No Gumbel-Sigmoid module  & $0.3293$ & $64.59\%$ \\
    \bottomrule
    \end{tabular}
\end{table}

\section{Conclusion}

This work presents a modular architecture based on a counterfactual entity-framing graph that effectively distinguishes critical narratives from conspiracy theories. By reasoning over intermediate graph representations, our framework applies counterfactual inference to uncover the underlying drivers of the model's classification decisions, achieving state-of-the-art efficacy with an $F_1$-score of 0.93. Three principles underpin this result. First, \textbf{domain adaptation} through pre-training on the extensive LOCO dataset substantially improved the model's ability to generalize, overcoming the limitations of the much smaller task-specific dataset. Second, \textbf{relational representation}: structuring entity relationships as graphs proved a superior way to encode the complex semantic information contained in the text. Third, \textbf{counterfactual interpretability}: these intermediate graph structures afford a transparent view into the model's decision-making process.

\paragraph{Future work.}
We identify two main directions for future research. First, more sophisticated methods are needed to mitigate spillover effects, with the goal of producing a more accurate and faithful counterfactual description. Second, exploring alternative approaches for the entity-framing phase, such as Deep SpERT \citep{zhu2023deepspanrepresentationsnamed}, could potentially lead to superior results in the initial entity extraction pipeline.

\section{Limitations}

Although our architecture achieves state-of-the-art performance, outperforming all other approaches on the target dataset, its cross-dataset generalization remains unverified. While we addressed local data scarcity through domain-adaptive pre-training on the larger LOCO corpus, we have not yet assessed how robust the framework is across different linguistic sources, since very little data with comprehensive annotations is currently available for highly specialized domains such as oppositional and conspiracy narratives.

A second limitation concerns the counterfactual component itself. The PEHE of 0.2994 indicates a non-negligible error in the estimation of individual treatment effects, so the recovered minimal subgraphs should be read as evidence about which entities the model relies on, not as a validated causal account of the narrative. The evaluation is also restricted to a single binary task in one language, and the leave-one-out supervision signal is derived from the classifier itself rather than from human-annotated counterfactuals.

\section{Broader Impact \& Ethical Considerations}

The capability to detect coordinated narratives cuts both ways. It can help identify and prevent disinformation campaigns targeting public health or coordinated attacks against democratic institutions. Nonetheless, it may also be misused as an automated censorship tool for detecting so-called ``undesired'' speech, enabling the identification and suppression of political opponents.

This dual-use risk is the reason we consider the interpretability component to be part of the contribution. A system that outputs only a binary label offers no recourse to a user whose message is flagged, whereas the minimal subgraph makes explicit which entities and interactions drove the decision, and can be inspected by a moderator, an auditor, or the affected author. We caution that the distinction between conspiratorial and merely critical speech is contested and context-dependent: the labels used here reflect the annotation decisions of the PAN corpus \citep{korenvcic2024overview} and are not a ground truth about legitimate speech. Any deployment should therefore act as decision support under human oversight, not as an autonomous moderation mechanism.

\begin{ack}
This work was supported by the Basque Government under grant DEUSTEK6 -- Humanized Computing for Smart Sustainable and Healthier Communities and Environments (IT1901-26), by the Ministry of Economy, Industry, and Competitiveness of Spain under Grant No.\ INCEPTION (PID2021-128969OB-I00), and by the European Union's Horizon Europe Research and Innovation Programme under the LATE-AYA project (Grant Agreement No.~101214326).

The authors declare no competing interests and no conflicts of interest.
\end{ack}

{
\small
\bibliographystyle{plainnat}
\bibliography{bibliography}
}

\clearpage
\appendix

\section{Span Enumeration Algorithm \& Hyperparameters}
\label{sec:appendix}

Algorithm~\ref{alg:hypergraph_vertices} describes the procedure we designed to extract entities from token representations using contextual embeddings and a max-pooling aggregation strategy. The algorithm is formulated under the span-nesting assumption stated in Section~\ref{sec:span}.

To ensure computational tractability, the algorithm constrains the maximum allowable length of each candidate span. This restriction reduces the space of all possible candidates to a manageable and computationally feasible set. Table~\ref{tab:hyperparameters} reports the full hyperparameter configuration used in all reported experiments.

\begin{algorithm}[h]
    \caption{Bipartite graph vertex selection}
    \label{alg:hypergraph_vertices}
    \begin{algorithmic}[1]
        \REQUIRE Embedding sequence $T = (h_0, \dots, h_{L-1})$ of length $L$, where $h_0$ corresponds to \texttt{[CLS]} and some $h_\ell$ corresponds to \texttt{[SEP]}; maximum span size $K$; $p$ embedding projection function; and $y$ span classification function.
        \ENSURE Matrix $E \in \mathbb{R}^{N \times d}$ containing the embeddings of the hypergraph vertices.

        \STATE $c \leftarrow T[0]$
        \STATE $i \leftarrow 1$
        \STATE $k \leftarrow K$
        \STATE $E \leftarrow [\ ]$ \COMMENT{(empty) list of selected embeddings}

        \WHILE{$k \geq 1$}
            \IF{$i = L$}
                \STATE \textbf{return} $E$
            \ENDIF

            \STATE $w \leftarrow W(k)$
            \STATE $s \leftarrow T[i : i + k]$

            \IF{$y(s, c, w) \notin \{\textsc{None}\}$}
                \STATE add $p(s, c)$ to $E$
                \IF{$L - (i + k) < k$}
                    \STATE $k \leftarrow L - (i + k)$
                    \STATE $i \leftarrow L - k$
                \ELSIF{$|E| = 0 \wedge K > 1$}
                    \STATE $k \leftarrow k - 1$
                    \STATE $i \leftarrow 0$
                \ELSE
                    \STATE $i \leftarrow i + k$
                \ENDIF

            \ELSE
                \IF{$i + k \geq L$}
                    \STATE $k \leftarrow L - (i + 1)$
                \ELSE
                    \STATE $i \leftarrow i + 1$
                \ENDIF
            \ENDIF
        \ENDWHILE

        \STATE \textbf{return} $E$
    \end{algorithmic}
\end{algorithm}

\begin{table}[h]
\caption{Hyperparameters of the experiment and their roles in the architecture. Note that the hyperparameter \texttt{num\_hyperlink} refers to the set size $k$, and all hyperlink references correspond to the set $E$.}
\label{tab:hyperparameters}
\centering
\small
\begin{tabular}{lll}
\toprule
\textbf{Parameter} & \textbf{Value} & \textbf{Description} \\
\midrule
\multicolumn{3}{c}{\textit{Span and entity representation}} \\
\midrule
max\_span\_width & 5 & Maximum number of tokens allowed in a candidate span. \\
num\_span\_type & 4 & Number of span/entity categories considered by the model. \\
entity\_dim & 16 & Dimensionality to represent each entity. \\
dropout\_rate & 0.15 & Dropout probability applied for regularization. \\

\midrule
\multicolumn{3}{c}{\textit{Hyperlink module}} \\
\midrule
num\_hyperlink & 7 & Maximum number of hyperlink types or relation channels. \\
hyperlink\_features\_dim & 16 & Dimensionality of hyperlink feature representations. \\
hyper\_link\_classifier\_hidden\_dim & [] & Hidden layer sizes (empty = linear classifier). \\
hyper\_link\_classifier\_activation & ELU & Activation function used in the hyperlink classifier. \\
gumbel\_sigmoid\_tau & 0.5 $\rightarrow$ 0 & Temperature annealing schedule for Gumbel-Sigmoid. \\
gumbel\_sigmoid\_hard & True & Uses hard/discrete samples during the forward pass. \\

\midrule
\multicolumn{3}{c}{\textit{Enricher (Transformer)}} \\
\midrule
enricher\_n\_heads & 4 & Number of attention heads in the Transformer enricher. \\
enricher\_num\_layers & 1 & Number of Transformer encoder layers. \\
enricher\_dim\_feedforward & 32 & Hidden dimension of the feed-forward network. \\
enricher\_use\_cls & True & Uses the [CLS] representation as a global context vector. \\
enricher\_max\_len & 200 & Maximum input sequence length. \\
enricher\_padding\_value & 0.0 & Padding value used for sequence batching. \\

\midrule
\multicolumn{3}{c}{\textit{Hypergraph backbone}} \\
\midrule
hyper\_graph\_feature\_extractor & HGT & Hypergraph encoder based on an HGT. \\
hyper\_graph\_hidden\_dim & 64 & Hidden feature dimension of graph layers. \\
hyper\_graph\_out\_dim & 64 & Output feature dimension produced by the graph encoder. \\
hyper\_graph\_heads & 4 & Number of attention heads in HGT layers. \\
hyper\_graph\_dropout & 0.15 & Dropout applied within the hypergraph encoder. \\
hgt\_num\_layers & 2 & Number of stacked HGT layers. \\
num\_outputs & 1 & Number of output prediction targets. \\

\midrule
\multicolumn{3}{c}{\textit{Training setup}} \\
\midrule
Optimizer & AdamW & Optimization algorithm used for training. \\
Weight decay & $10^{-5}$ & L2 regularization coefficient. \\
Learning rate & $10^{-4}$ & Initial learning rate. \\
Batch size & 500 & Number of samples processed per optimization step. \\
Epochs & 10 & Number of full passes through the training dataset. \\
\bottomrule
\end{tabular}
\end{table}

\end{document}